\documentclass[sigconf]{acmart}

\usepackage{algorithm}
\usepackage{algorithmic}
\usepackage{amsmath}
\AtBeginDocument{%
  \providecommand\BibTeX{{%
    \normalfont B\kern-0.5em{\scshape i\kern-0.25em b}\kern-0.8em\TeX}}}

\setcopyright{acmcopyright}
\copyrightyear{2022} 
\acmYear{2022} 
\setcopyright{rightsretained} 
\acmConference[KDD '22]{Proceedings of the 28th ACM SIGKDD Conference on Knowledge Discovery and Data Mining}{August 14--18, 2022}{Washington, DC, USA}
\acmBooktitle{Proceedings of the 28th ACM SIGKDD Conference on Knowledge Discovery and Data Mining (KDD '22), August 14--18, 2022, Washington, DC, USA}
\acmDOI{10.1145/3534678.3539171}
\acmISBN{978-1-4503-9385-0/22/08}

\usepackage{etoolbox}
\makeatletter
    \patchcmd{\maketitle}{\@copyrightpermission}{
   \begin{minipage}{0.3\columnwidth}
     \href{https://creativecommons.org/licenses/by/4.0/}{\includegraphics[width=0.90\textwidth]{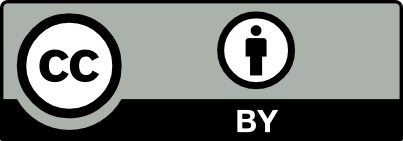}}
   \end{minipage}\hfill
   \begin{minipage}{0.7\columnwidth}
     \href{https://creativecommons.org/licenses/by/4.0/}{This work is licensed under a Creative Commons Attribution International 4.0 License.}
   \end{minipage}
  
   \vspace{5pt}
}{}{}

\makeatother

\begin{document}

%%
%% The "title" command has an optional parameter,
%% allowing the author to define a "short title" to be used in page headers.
\title{Automatic Controllable Product Copywriting for E-Commerce}

%%
%% The "author" command and its associated commands are used to define
%% the authors and their affiliations.
%% Of note is the shared affiliation of the first two authors, and the
%% "authornote" and "authornotemark" commands
%% used to denote shared contribution to the research.
%\author{Ben Trovato}
%\authornote{Both authors contributed equally to this research.}
%\email{trovato@corporation.com}
%\orcid{1234-5678-9012}
\author{Xiaojie Guo}
%\authornotemark[1]
\email{xguo7@gmu.edu}
\affiliation{
  \institution{JD.COM Silicon Valley Research Center}
%  \streetaddress{P.O. Box 1212}
%  \city{Dublin}
%  \state{Ohio}
  \country{USA}
%  \postcode{43017-6221}
}

\author{Qingkai Zeng}
\authornote{Part of the work was done when QK.Zeng was an intern at JD.COM Silicon Valley Research Center.}
\email{qzeng@nd.edu}
\affiliation{
  \institution{University of Notre Dame}
%  \streetaddress{P.O. Box 1212}
%  \city{Dublin}
%  \state{Ohio}
  \country{USA}
%  \postcode{43017-6221}
}

\author{Meng Jiang}
\email{mjiang2@nd.edu}
\affiliation{
  \institution{University of Notre Dame}
%  \streetaddress{P.O. Box 1212}
%  \city{Dublin}
%  \state{Ohio}
  \country{USA}
%  \postcode{43017-6221}
}
%\authornote{JD.COM Silicon Research Valley Center}
\author{Yun Xiao}
%\authornotemark[1]
\email{xiaoyun1@jd.com}
\affiliation{
  \institution{JD.COM Silicon Valley Research Center}
%  \streetaddress{P.O. Box 1212}
%  \city{Dublin}
%  \state{Ohio}
  \country{USA}
%  \postcode{43017-6221}
}

%\authornote{JD.COM Silicon Valley Research Center}
\author{Bo Long}
%\authornotemark[1]
\email{bo.long@jd.com}
\affiliation{
  \institution{JD.COM}
%  \streetaddress{P.O. Box 1212}
%  \city{Dublin}
%  \state{Ohio}
  \country{China}
%  \postcode{43017-6221}
}

%\authornote{JD.COM Silicon Research Valley Center}
\author{Lingfei Wu}
%\authornotemark[1]
\email{lwu@email.wm.edu}
\affiliation{
  \institution{JD.COM Silicon Valley Research Center}
%  \streetaddress{P.O. Box 1212}
%  \city{Dublin}
%  \state{Ohio}
  \country{USA}
%  \postcode{43017-6221}
}

\renewcommand{\shortauthors}{Xiaojie Guo et al.}
%%
%% By default, the full list of authors will be used in the page
%% headers. Often, this list is too long, and will overlap
%% other information printed in the page headers. This command allows
%% the author to define a more concise list
%% of authors' names for this purpose.
%\renewcommand{\shortauthors}{Trovato and Tobin, et al.}

%%
%% The abstract is a short summary of the work to be presented in the
%% article.
\begin{abstract}
Automatic product description generation for e-commerce has witnessed significant advancement in the past decade. Product copywriting aims to attract users’ interest and improve user experience by highlighting product characteristics with textual descriptions. As the services provided by e-commerce platforms become diverse, it is necessary to adapt the patterns of automatically-generated descriptions dynamically.
In this paper, we report our experience in deploying an E-commerce Prefix-based Controllable Copywriting Generation (EPCCG) system into the JD.com e-commerce product recommendation platform. The development of the system contains two main components: 1) copywriting aspect extraction; 2) weakly supervised aspect labelling; 3) text generation with a prefix-based language model; and 4) copywriting quality control. We conduct experiments to validate the effectiveness of the proposed EPCCG. In addition, we introduce the deployed architecture which cooperates the EPCCG into the real-time JD.com e-commerce recommendation platform and the significant payoff since deployment. The codes for implementation are provided at \url{https://github.com/xguo7/Automatic-Controllable-Product-Copywriting-for-E-Commerce.git}.
%The ACPCG system has been deployed in JD.com since Feb 2021. By December 2021, it has generated --- million product descriptions, and improved the overall averaged click-through rate (CTR) and the Conversion Rate (CVR) by 4.22% and 3.61%, compared to base- lines, respectively on a year-on-year basis. The accumulated Gross Merchandise Volume (GMV) made by our system is improved by 213.42%, compared to the number in Feb 2021.
\end{abstract}

%%
%% The code below is generated by the tool at http://dl.acm.org/ccs.cfm.
%% Please copy and paste the code instead of the example below.
%%


%%
%% Keywords. The author(s) should pick words that accurately describe
%% the work being presented. Separate the keywords with commas.

%% A "teaser" image appears between the author and affiliation
%% information and the body of the document, and typically spans the
%% page.

%%
%% This command processes the author and affiliation and title
%% information and builds the first part of the formatted document.
\maketitle

\section{Introduction}

Informative product copywriting is critical for providing a desirable user experience on an e-commerce platform. Different from brick-and-mortar stores where salespersons can hold face-to-face conversations with customers, e-commerce stores heavily rely on textual and pictorial product descriptions which provide product information and eventually promote purchases~\citep{guo2021intelligent}. Accurate and attractive product descriptions help customers make informed decisions and help sellers promote products. 
% in the %efficiency of human copywriters and huge manpower cost.
% cannot
 Traditionally, human copywriters perform product copywriting, which exposes significant limitations to match the growth rate of new products.
 
\begin{figure}[htb]
\centering
\includegraphics[width=1.0\columnwidth]{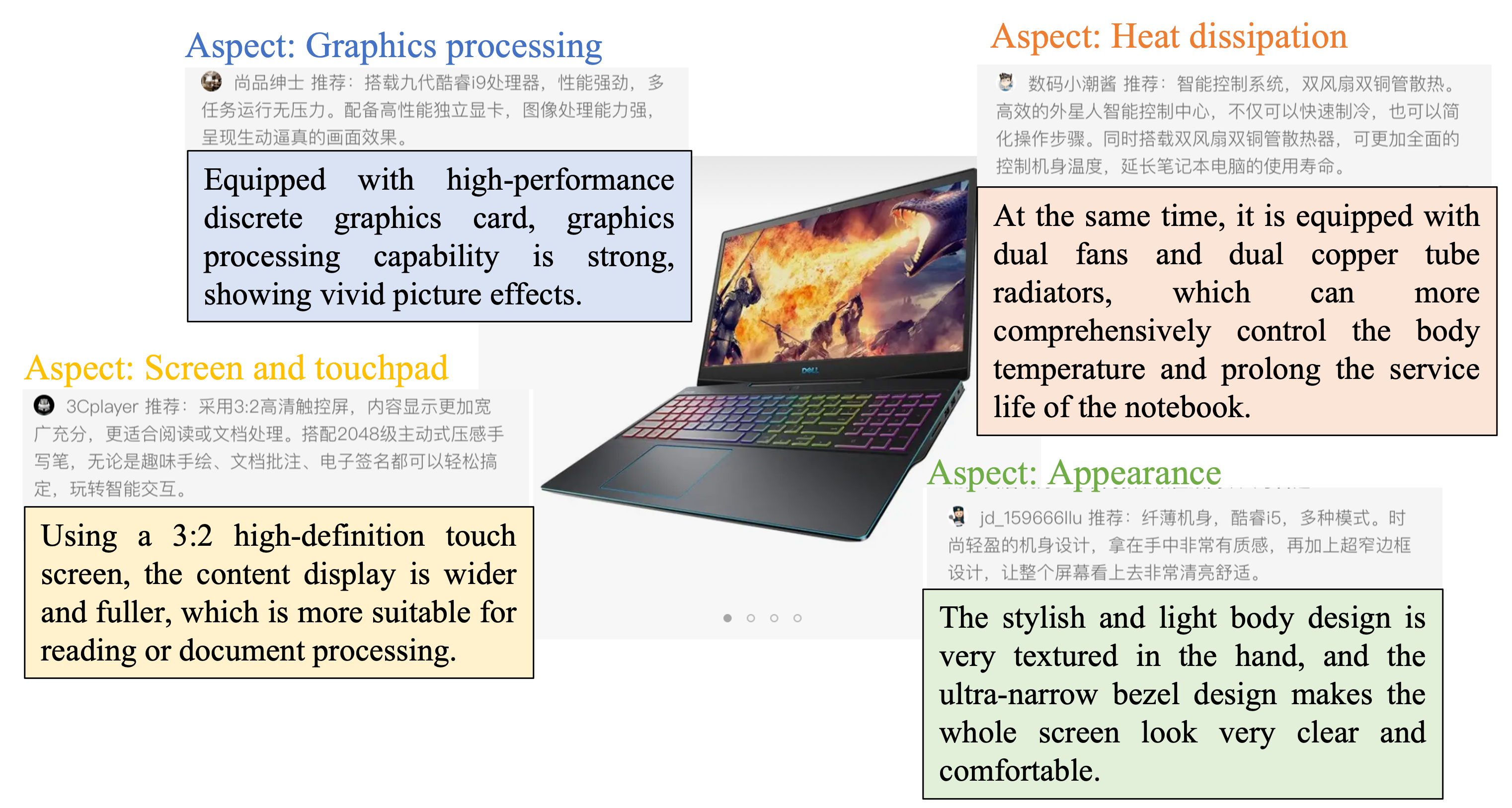} \vspace{-0.3cm}
\caption{\small{Examples of diverse product copywriting with various aspects: in the example of copywriting for computers, there are aspects related to appearances, graphics processing, screen, touch-pad, heat dissipation, etc.} }\vspace{-0.3cm}
\label{fig:example}
\end{figure}

% Second,  increases as the number of products increases. 
%Last but not least, specialist training and tutoring for copywriters for different products and scenarios are expensive~\citep{zhang2021automatic}. 
To address these issues, automatic product copywriting generation has become an essential line of research in e-commerce. 
%Thanks to the Natural language generation (i.e., NLG) techniques, it is possible to produce understandable texts in human languages from some underlying non-linguistic or textual representation of information~\citep{chen2019towards, langkilde1998generation}.
Due to the great success of seq2seq models on natural language generation tasks, researchers adopt various neural architectures \cite{sutskever2014sequence, raffel2020exploring,vaswani2017attention,vinyals2015pointer} in this framework
to utilize both the user-generated reviews and product information for product description/copywriting generation. These works provide the inspiration for designing the specific generators in generating product descriptions for e-commerce products~\citep{chen2019towards,wang2017statistical, daultani2019unsupervised,lipton2015capturing,xiao2019text,khatri2018abstractive}. 
%Daultani, Nio, and Chung (2019) propose generating extractive summaries for product descriptions based on a coverage maximization algorithm. Khatri, Singh, and Parikh (2018) introduce a document-context-based sequence-to-sequence (seq2seq) model to produce summaries for product descriptions. 

However, due to the diversity of e-commerce platform services, it is necessary to adapt the patterns of description writing to meet the different preferences of customers and sellers. As shown in Figure~\ref{fig:example}, various copywriting can show diverse aspects of the same product. Controlling the content aspects of the generated copywriting is significantly important for several reasons: (1) it helps attract different groups of customers with diverse copywriting contents. 
Different customers have different concerns about the same product. In other words, they will be attracted by most relevant aspect of the product based on their needs. 
%Diffrent people care about different aspects of a product. 
For example, business people may care about the duration of battery power of a mobile phone while the game fans pay more attention to the screen refreshing rate; (2) it is flexible for sellers to display the most attractive aspects of the products, especially when the recommendation is combined with special promotion event or activity. For example, 
on Father's day or Mother's day, highlighting the ``big screen for easy reading" and ``concise interface" of mobile phones help promote the mobile phone as gifts for parents. Thus, controlling the aspects of the generated copyrighting is of great importance towards the various requirements from both the customers and platform sellers.

%先说下controllable text generation技术的发展，尤其是content/aspect controllable. 再去e-commerce很少做到，提一下chen的work。然后提出工业界product decsription实现的难度。引出我们的模型方法以及应用结果。
With the development of controllable text generation technique, controllable product description generation becomes possible. \citet{chen2019towards} explored to generate personalized product descriptions controllable by the customer preferences. \citet{li2020aspect} and \citet{liang2021custom} presented an abstractive summarization system, where the summary can capture the most attractive aspects of a product. However, the methods mentioned above have several limitations for real-world controllable e-commerce product copywriting generation: (1) the generated copywriting from the generation models are likely to disrespect the truth of products, which is not acceptable to the real-world platform; (2) the aspects extracted by existing methods are based on adjective words in each sentence, without considering the sentence-level semantics; (3) existing clustering-based aspect assignment algorithm is hard to further involve more newly coming products for future model upgrading, which is impractical in many real-world scenarios; (4) there is no evaluation method to judge whether the generated copywriting matches the desired aspects. 

To solve the aforementioned challenges and more importantly, to successfully implement and deploy the controllable product copywriting generator in the real world, large-scale e-commerce platform, we propose the E-Commerce Prefix-based Controllable Product Copywriting Generation (EPCCG). In this paper, we report our experience in developing the EPCCG system and deploying the EPCCG into the JD.com e-commerce product recommendation platform. The development of EPCCG consists of three main steps: 1) aspect extraction from copywriting, which is based on latent Dirichlet allocation (LDA) to extract aspects from copywriting corpus; (2) a novel phrase-based aspect classifier to label the copywriting as training data preparation; (3) product copywriting generation, which is built from a prefix-based e-commerce pre-trained model; and 4) knowledge-based post-processing, To the best of our knowledge, the proposed EPCCG is the first successful deployed controllable product copywriting generation system in the real-word e-commerce platform. Our contributions are summarized as follows:\vspace{-0.2cm}

\begin{enumerate}
    \item A domain-specific generative model EPCCG for product copywriting generation is proposed and deployed for the JD.com e-commerce product recommendation platform.
  \item We propose a phrase-based aspect classification method for automatically labelling the product copywriting with the extracted aspect. It can also be used for evaluating the aspect capturing ability of EPCCG. 
  \item We extend the basic EPCCG model to Prompt-EPCCG by exploring various prompt strategies. The Prompt-EPCCG shows significant improvement towards the quality of generated product copywritng. 
\item We introduce the overall architecture of deployed system where the EPCCG is implemented into the large-scale JD.com recommendation platform. The experience learnt during deployment is also summarized.
\item The experimental exploration results and the significant payoff since deployment demonstrate the superiority of the proposed method over the baseline models and the effectiveness of its deployment in a real-world scenario. 
\end{enumerate}

\section{Related work}

%\subsection{}
%\subsubsection{Text generation}
%The typical technqiue based on neural network for text generation is the attention-based Seq2Seq model~\citep{sutskever2014sequence, raffel2020exploring, vaswani2017attention, vinyals2015pointer}. The attention-based Seq2Seq model has demonstrated its effectiveness in a number of tasks of text generation, including neural machine translation~\citep{wu2016google,sutskever2014sequence}, abstractive text summarization~\citep{nallapati2016abstractive}, dialogue generation~\citep{serban2017multiresolution}, etc. Generally speaking, the Seq2Seq model has become one of the most common frameworks for text generation. Transformer, as an important component for semantic embedding, has achieved state-of-the-art results in neural machine translation and rapidly become a popular technique for sequence-to-sequence learning due to its outstanding performance and high efficiency. Based on this, \citet{devlin2018bert} proposed BERT, a pre-trained language model, has achieved the significant performances on natural language processing tasks.

\textbf{Controllable text generation in Academic Exploration}
The typical technique based on neural network for text generation is the attention-based Seq2Seq model~\citep{sutskever2014sequence, raffel2020exploring, vaswani2017attention, vinyals2015pointer,wu2018word}. The attention-based Seq2Seq model has demonstrated its effectiveness in a number of tasks of text generation, including neural machine translation~\citep{sutskever2014sequence}, abstractive text summarization~\citep{nallapati2016abstractive}, dialogue generation~\citep{serban2017multiresolution}, etc.
 Controllable Text Generation (CTG) has recently attracted more focus from many researchers in the NLP community. The most relevant sub-topic in CTG to our problem is the Topic-based Generation\citep{khalifa2020distributional, keskar2019ctrl, dathathri2019plug}. CTRL~\citep{keskar2019ctrl} is an early attempt in this direction which trains a language model (i.e., Transformer) conditioned on a variety of control code including domain, style, topics, dates, entities, relationships between entities, etc.
 %\cite{khalifa2020distributional} explored the distributional view for controlled text generation formalized as a constraint satisfaction problem over the probability distribution of the desired target language model. To make the generated text more coherent to the topics, \citet{tang2019topic} adapts the encoder of the topic modeling component for a discriminator. 
 Though great efforts have been put into the exploration of controllable text generation, extending them into real-world product copywriting generation still remain large challenges, such as inconsistent attribute information before generated text and product information, and unavailability to the labeled copywriting datasets.

%\subsubsection{Name label classification}
\begin{figure*}[htb]
\centering
\includegraphics[width=0.98\textwidth]{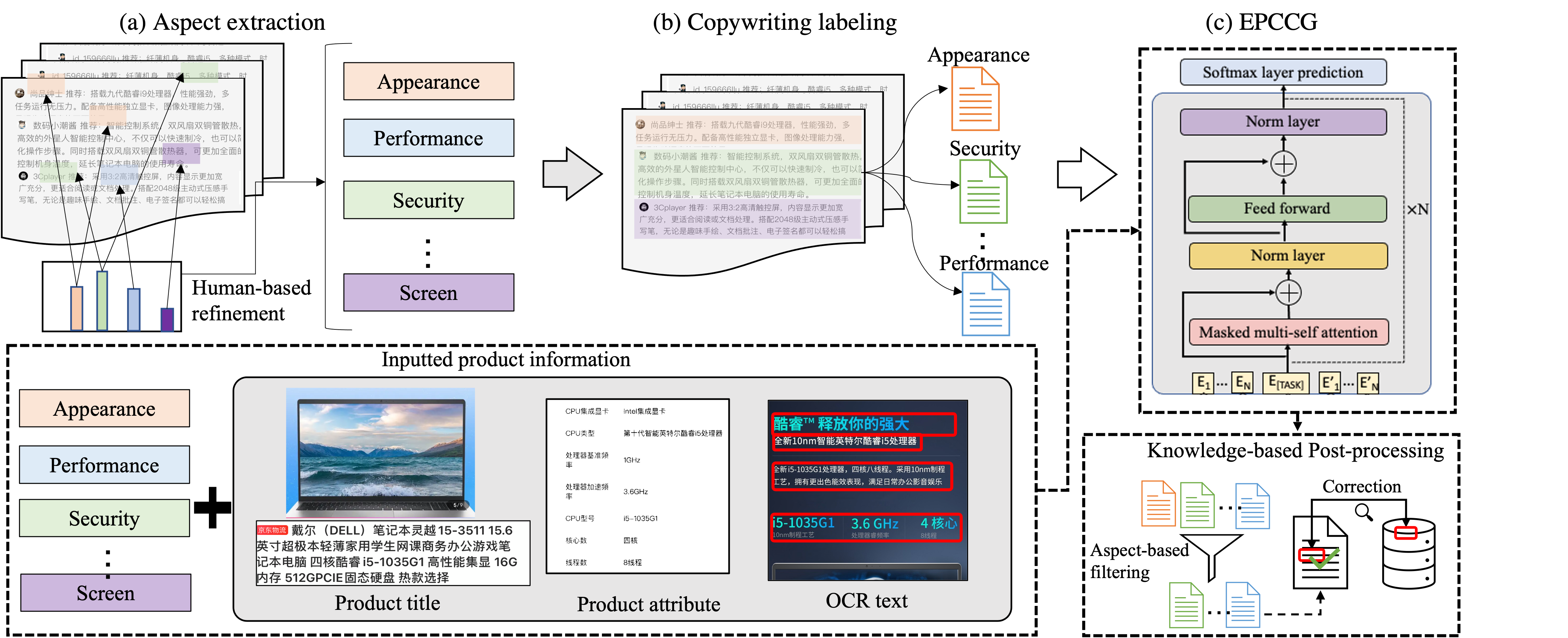} 
\caption{\small{The development workflow of the proposed EPCCG: Training process: First, the aspects of a given kind of category of products are extracted (see sub-figure (a)), and the collected copywriting training samples are labeled with aspects (see sub-figure (b)); then the labeled copywriting are used for training the EPCCG (see sub-figure (c)). Inference/Generating process (circled in black dotted line): For each product, the product information and each desired aspect are combined as input, following which the knowledge-based post-process is conducted to guarantee the quality of the generated copywriting further.}}
\label{fig:frame}
\end{figure*}

\textbf{Product description generation}
Previous studies for text description generation in the e-commerce domain focused on statistical frameworks such as\citep{wang2017statistical, gerani2014abstractive,xiao2019text,zhang2021automatic}, which incorporate statistical methods with the template for product descriptions generation. %\cite{gerani2014abstractive} also generates summarization of product reviews by applying a template-based NLG framework. 
Such methods are limited by the hand-crafted templates. To this end, researchers adopted deep learning models and introduced diverse conditions into the generation model. \citet{lipton2015capturing} generated reviews based on the conditions of semantic information and sentiment by language model. \citet{khatri2018abstractive} proposed a novel Document-Context based Seq2Seq models for abstractive and extractive summarizations in e-commerce. %\citet{li2020aspect} propose a multimodal summarizer to improve the importance, non-redundancy, and readability of the product summarization. 
All aforementioned methods are for general product description generation without controlling the aspects of the generated contents.

\textbf{Controllable Product description generation}
To meet the diverse pattern requirement of the generated product description, limited number of attempts have been made to control the generated product description regarding the length, sentiments and topics. \citet{chen2019towards} explored to generate personalized product descriptions controllable by the customer preferences and adjective topics. \citet{li2020aspect, liang2021custom} presented an abstractive summarization system that produces summary for Chinese e-commerce products, where the summary can capture the most attractive aspects of a product. However, these works all focus on introducing the single product generation model without matured deployment exploration in the large-scale e-commerce platform, where (1) an efficient and reusable aspect extraction and automatic data labeling process and (2) the precise consistent between copywriting and product information are required.  

\section{EPCCG: The Proposed Technology}

The development of EPCCG consists of three main parts, as illustrated in Figure~\ref{fig:frame}. The first step is aspect extraction where the main aspects of each category of products are extracted from copywriting corpus. The second step is about copywriting labeling, namely, classifying the unlabeled copywriting samples based on the extracted aspects. In the third step, given the labeled copywriting, the E-commerce prefix-based controllable copywriting generator (EPCCG) is well trained and utilized. During production, the generated copywriting from EPCCG will be inputted into a correction and filtering module to improve their quality to meet the requirement of the industry e-commerce platform. The details of each main step are provided as follows:

\subsection{LDA-based Aspect Extraction}
\label{sec:aspect ext}
Given the large volumes of product copywriting of a category of products, such as ``skin care''  or ``mobile phones'', aspect extraction aims to extract the key and popular topics from the copywriting written by professional human writers. Here we summarize the sub-problem formulation.

\paragraph{\textbf{Sub-problem formulation: aspect extraction}}
Given the collected copywriting datasets $\mathcal{D}=\{x_i\}^{N}_{i=1}$ from the human writers of a given product category, where $N$ is the number of copywriting samples. The goal is to extract the key aspects $\mathcal{T}=\{t_m\}_{m=1}^M$ of the given product category, where $M$ denotes the number of aspects. \vspace{0.1cm}

We proposed a LDA-based extractor for aspect extraction. The extractor contains two steps, as shown in Figure~\ref{fig:frame}(a). First, the initial aspects are extracted by LDA model. Second, human-based refinement is conducted to balance the training samples and guarantee that the extracted aspects meet the e-commerce market preferences. We elaborate on each step as follows.

\subsubsection{Topic Modeling via LDA}
Topic Modeling in natural language documents aims to use unsupervised learning to extract the main topics in a collection of documents. 
LDA~\citep{jelodar2019latent} is an unsupervised generative probabilistic method for modeling the semantic topic in corpus, which is the most commonly used topic modeling method. It assumes that each document can be represented as a probabilistic distribution over latent topics, and that topic distribution in all documents share a common Dirichlet prior. Each latent topic in the LDA model is also represented as a probabilistic distribution over words and the word distributions of topics share a common Dirichlet prior as well. Due to the space limit, more details of the typical LDA algorithm are provided in Appendix. \ref{app:aspect}.

%In the process of product aspect extraction, given the copywriting dataset $\mathcal{D}=\{x_i\}^{N}_{i=1}$ and each copywriting $x_i$ contains $H^{(i)}$ words $w_k^{(i)}$.
%the whole copywriting corpus $\mathcal{D}$ is modeled as the following process:
%  \begin{itemize}
%      \item  Choose a distribution $\phi \backsim Dir(\beta)$ for aspect.
%      \item Choose a distribution $\theta \backsim Dir(\alpha)$ for copywriting.
%      \item For each word $w_k^{(i)}$ in copywriting $x_i$:\\
%          (a) Choose a topic $t_m^{(i)} \backsim Multinomial(\phi)$.\\
%          (b) Choose a word $w_k^{(i)}\backsim Multinomial(\theta)$.
%  \end{itemize}
%to learn this generative model, we maximize the objective as:
%\begin{equation}\nonumber
%    \max_{\phi, \theta} \prod^N_{i=1}\int p(\theta^{(i)}|\alpha)(\prod^{H^{(i)}}_{k=1}\sum_{t_m^{(i)}}p(z_m^{(i)}|\theta^{(i)})p(w_k^{(i)}|t_m^{(i)},\beta)))d\theta^{(i)},
%\end{equation}
%where $\alpha$, $\beta$ and $M$ are hyper-parameters about LDA. %Here we use the Gibbs sampling, which is a Monte Carlo Markov-chain algorithm to estimate the LDA parameters. After this, we can successfully identify the main aspects from the copywriting corpus.
To select the best-performer hyper-parameters, we propose a dynamic-LDA inference-process. Specifically, we adopt the aspect coherence score to evaluate the performance of the LDA under each hyper-parameter setting. Aspect coherence measures the degree of semantic similarity between each pair of high-scoring words in each aspect. We then select the LDA model which has the highest coherence score. Given the well-learned LDA model, we can select the aspects which have the most elevated aspect distribution probability. In this way, we can get the aspects of the given category of product $\mathcal{T}=\{t_m\}^M_{m=1}$ where each aspect contains several keywords.

\subsubsection{Human-based refinement}
Based on the keywords of each extracted aspect, we could easily interpret the semantic meaning of each aspect. However, for real-world industry application, the initial extracted aspects from the LDA model have several limitations: (1) the amount of training samples assigned for each aspect is not balanced, mainly influencing the following training process of the controllable text generation model; (2) some extracted aspects are not popular to attract customers, while some well-recognized aspects by professional marketing sellers are not discovered. Thus, we propose involving human-based refinement to adjust the extracted aspects slightly Based on the modified aspects and their semantics, it is easy to come up with the name for each aspect\footnote{For brevity, we also use $t_i$ to refer the name of each aspect.}. 

\subsection{Unsupervised Phrase-based Aspect classification}
\label{sec:aspect_classify}

A severe challenge of copywriting generation while real-world deployment is the lack of aspect-labeled copywriting. In this section, we propose the phrase-based aspect classifier, which can be utilized to assign the aspect to each product copywriting without human efforts. In addition, this well-trained aspect classifier can be utilized further to evaluate whether each generated copywriting match the desired aspect.

\textbf{Sub-problem formulation: Extremely Weakly Supervised Aspect
classification}. Given the collected copywriting datasets $\mathcal{D}=\{x_i\}^{N}_{i=1}$ as well as the names of extracted key aspects $\mathcal{T}=\{t_m\}_{m=1}^M$ of a given product category, where $M$ denotes to the number of aspects and $N$ denotes to the number of copywriting, the goal is to assign the aspect to each copywriting as $f:x_i\longrightarrow t_i$.

 %In the following, we first review the preliminary of name label classifier (i.e.,\textit{LOTClass}) and its limitations, and then introduce the proposed novel phrase-based LOTClass for Chinese document. 

\subsubsection{Limitations of existing LOTClass}

We proposed a phrase-based aspect classifier inspired by the \textit{LOTClass}~\citep{meng2020text} for aspect classification using only the names of aspects. The goal of \textit{LOTClass} is only to use the label name of each class to train a classifier on unlabeled data without using any labeled documents. The main idea is to utilize the pre-trained language model for category name understanding to generate pseudo label for classification task and fine-tuning classifier on generated pseudo label data.
However, the existing LOTClass has several limitations in dealing with the Chinese e-commerce product copywriting classification: (1) \textbf{Different language characteristic between Chinese and English}. Compared with English, Chinese single characters cannot carry out effective semantic information, and thus phrases are needed to better describe the semantic information of category names. 
%However, the LOTClass and most of the existing Chinese pre-training models are based on characters not phrases. 
(2) \textbf{Limited expressive ability in e-commerce domain}. 
%The number of Chinese phrases is significantly larger than the number of single words, and 
The general phrase vocabulary cannot have limited coverage for the specific domain, e.g., e-commerce. 
%Thus the expressive ability of the existing pre-training language model for downstream task in a specific field is limited. 
Therefore, we have proposed the novel phrase-based LOTClass for e-commerce in Chinese to solve the above limitations. 

\subsubsection{The Overall workflow of phrase-based LOTClass}

The phrase-based LOTClass consists of three main steps, as shown in Algorithm\ref{alg}. 
First and the most important, we need to train a domain specific phrase-based pre-trained language model $P(x)$. The details of training a phrase-based pre-trained language model is provided in the following Section~\ref{sec:phrase-based lm}. 
Second, after getting the e-commerce pre-trained language model, we rely on it to find a set of semantically similar substitute words for the name of the aspect To do this, in the corpus, e.g., copywriting dataset, for each aspect $t_m$, we mask the aspect and utilize the Phrase-based  pre-training model to make predictions for these masked position. Normally, we select the top-$K$ words based on their appearance frequency in the overall prediction to get the semantically similar substitute words set $\mathcal{S}^{(m)}$ for each aspect. 
At last, each copywriting is classified based on its coverage over the substitute words of the aspects. Specifically, as shown in Algorithm\ref{alg}, we calculate the coverage number $c_i^{(m)}$ of a given copywriting $x_i$ regarding the substitute words set $\mathcal{S}^{(m)}$ of each aspect. The aspect that has the largest $c_i^{(m)}$ will be assigned to the copywriting $x_i$. 
As a result, all the copywriting will be labeled with an aspect.

\begin{algorithm}[htb]
\caption{Phrase-based LOTClass}
\label{alg:A}
\begin{algorithmic}
\REQUIRE The collected copywriting datasets $\mathcal{D}=\{x_i\}^{N}_{i=1}$; The names of extracted key aspects $T=\{t_m\}_{m=1}^M$; The phrase-based pre-trained language model $P(x)$.
\ENSURE The substitute words set $\mathcal{S}^{(m)}$ for each aspect $t_m$; The assigned aspect $y_i \in \mathcal{T}$ for each product copywriting $x_i$.\\
\FOR{$m=1:M$}
\STATE  Mask the aspect $t_m$ in all of the copywriting of the corpus $\mathcal{D}$ to make the masked document $\bar{\mathcal{D}}$.
\STATE  Input the masked document $\bar{\mathcal{D}}$ to the pretrained language model $P(x)$ to predict the masked words. 
\STATE  Based on the predicted words set $\bar{\mathcal{S}}^{(m)}$, select the Top-$K$ words with the highest frequent appearances to form the substitute words set $\mathcal{S}^{(m)}$.  \\
\ENDFOR
\FOR{$i=1:N$}
\FOR{$m=1:M$}
\STATE calculate the coverage $c_i^{(m)}$ of $x_i$ regarding the substitute words set $\mathcal{S}^{(m)}$
\ENDFOR
\STATE Classify the copywriting $x_i$ as $y_i=t_m$ where $c_i^{(m)}$ has the highest value.
\ENDFOR
\STATE 
\end{algorithmic}
\label{alg}
\end{algorithm}

%\paragraph{\textbf{A.Find the substitute words for label name}. Our goal in the first step is to find a set of semantically similar substitute words for the label names. To do this, in the corpus data, e.g., copywriting, we mark the label names and utilize the Phrase-based Chinese Pre-training mode to make predictions for these masked positions. In this way, each label name can has a set of predicted semantically similar substitute words. Normally, we select the top-$k$ strong-related works based on their appearance frequency in the whole corpora.}

%\paragraph{\textbf{B. Classifying the copywriting} To train a classifier, the key point is to automatically label the unlabeled datasets. To achieve this, we resort to the above substitute words for each label name. Specifically, for a copywriting, if it contains more than $N$ substitute words of a label name, it be classified to the class of the label name. As shown in Figure 3 (b). Based on this methods, all the copywriting will be labeled with a label name, i.e., each copywriting will be assigned with a topic.}

\subsubsection{Phrase-based Pre-trained Language Model}
\label{sec:phrase-based lm}

The core of the phrase-based aspect classifier is to learn a powerful e-commerce phrase-based pre-trained language model. 
% The existing pre-trained model is either character-based for English or cover the general knowledge. 
There are two major limitations of the existing pre-trained model: (1) the basic unit in the vocabulary set of the existing pre-trained model is token (character in Chinese and word in English) and is hard to be used in tasks that need phrase-level knowledge; (2) training on the corpus in general domain made it be lack of domain-specific knowledge.
Thus, we propose to train a e-commerce phrase-based pre-trained language model. 
%This step contains two main components: first is about building the domain specific phrase-based Chinese e-commerce vocabulary; Second is about fine-tuning the general language models with the extracted phrase-based Chinese e-commerce vocabulary.

\paragraph{\textbf{Step 1: Building phrase-based Chinese e-commerce vocabulary}}
The proposed phrase-based Chinese e-commerce vocabulary consists of general Chinese characters and extracted domain-specific phrases. The general Chinese characters can be obtained any open-source basic Chinese character vocabulary. To extract the domain specific Chinese phrases, specifically, we first use part-of-speech tagging method~\citep{brill1992simple} to extract high-frequency noun phrases from e-commerce corpus data as seed vocabulary. Since the number of these extracted phrases is limited and can not cover many insightful phrases. To make extension, the obtained seed vocabulary is used as the input of the phrase mining algorithm AutoPhrase~\citep{shang2018automated} which are performed on the existing copywriting corpus to get the domain specific phrases. %Finally, the extraction results $V_{domain}$ are merged into the initial general Chinese vocab list as $V=\{V_{init},V_{domain}\}$.

\paragraph{\textbf{Step 2: Phrase-based Chinese Pre-trained model}}
The next step is to tokenize the copywriting datasets based on the vocabulary. The principal of tokenization is to first find out all the phrases in the sentences which matches the phrase-based Chinese e-commerce vocabulary. The remaining sub-sentences will then be tokenized in a character-based fashion. The model architecture adopted here can be any language model.\footnote{For our deployment, we utilize the WoBERT\citep{zhuiyiwobert} as the initial model.}.

\subsection{Prefix-based Controllable Copywriting Generation}
In this section, we propose the E-commerce Prefix-based controllable copywriting generation model (EPCCG) and its extension, the Prompt-EPCCG. We first define the sub-problem for this step. Then, we introduce the preliminary of the basic prefix language model and the motivation to choose it. Next, we present the detailed techniques of the proposed EPCCG. Furthermore, we introduce the Prompt-EPCCG for the situation that only few labeled data can be used in training.

\textbf{Sub-problem formulation: controllable copywriting generation}
%Given product information sequences of the form $x = (x_1, . . . , x_n)$ where each $x_i$ comes from a fixed set of symbols, the goal of language modeling is to learn $p(x)$. Because $x$ is a sequence, it is natural to factorize this distribution using the chain rule of probability~\citep{keskar2019ctrl,bengio2003neural} as:
% \begin{equation}
%     p(x)=\prod^{n}_{i=1} p(x_i|x_{<i}),
% \end{equation}
%which decomposes language modeling into next-word prediction. While
Given product information text which is expressed in a sequences of the form $I = (w^I_1, . . . , w^I_{L_1})$ as well its relevant copywriting text $X = (w^X_1, . . . , w^X_{L_2})$, the goal is to learn a conditional language model $p_c(X)$ given the pre-defined aspect $y_i$ as:
 \begin{equation}
     p_c(X)=\prod\nolimits^{L_2}_{i=1} p(w^X_i|w^X_{<i},I,y_i),
     \label{eq:ctlr_dstr}
 \end{equation}
where aspect $y_i$ provides a point of control over the generation. 

\subsubsection{Prefix Language Model}

The prefix Language Model (LM)~\cite{dong2019unified} is a left-to-right Language Model that decodes output sentences on the condition of a prefixed input sequence, which is encoded by the same model parameters with a bi-directional mask. The preliminary knowledge of prefix LM can be referred in Appendix~\ref{sec:pre_LM}. 
%As shown in Figure 4, during training process, the input tokens can attend to each other bidirectionally, while the output tokens can only attend to tokens on the left. To realize this, a corrupted text reconstruction objective is usually applied over input tokens, and an auto regressive language modeling objective is applied over output tokens, which encourages the prefix LM to better learn representations of the input. In this way, the prefix LM can model the generative process of $p(w^X_i|w^X_{<i})$ for each sentences.

%Specifically, these input vectors is first packed into $H^0 = [y_i, w^I_1, . . . , w^I_{L_1}]$, and then encoded into contextual representations at different levels of abstract $H^l =[h^l_t,h^l_1,···,h^l_n]$ using an L-layer Transformer
%$H^l = Transformer^l (H^{l−1}), l \in [1, L]$. In each Transformer block, multiple self-attention heads are used to aggregate the output vectors of the previous layer. For the $l-th$ Transformer layer, the output of a self-attention head $A^l$ is computed via:

%where the previous layer’s output $H^{l−1} \in R^{n\times d_h}$ is linearly projected to a triple of queries, keys and values using parameter matrices $W^l_Q$ , $W^l_K$ , $W^l_V ∈ R^{d_h\times d_k}$, respectively, and the mask matrix $M \in R^{n\times n}$ determines whether a pair of tokens can be attended to each other.

The flexible model structure of prefix LM makes it suitable for our application, where the power of language is utilized to realize our special training goals. Language can specify different parts of inputs and outputs as a sequence of symbols flexibly. Multiple kinds of information (i.e., product information and controller aspects) are embedded into the inputs represented by languages with specific tricks to build up the dataset. The language model will learn the tasks implicitly. In this way, one model with the same structure can be reused for different tasks by only changing the input data which contains information of inputs,  outputs and tasks all together. 

\subsubsection{EPCCG: The Proposed Model}
To learn the conditional distribution in Eq~\ref{eq:ctlr_dstr} for E-commerce application, we proposed the E-commerce Prefix-based controllable copywriting generation model (EPCCG) based on the prefix LM. We utilize a 12-layer Transformer as our backbone network, where the input vectors are encoded to contextual representations through the Transformer blocks. Each layer of Transformer block contains multiple self-attention heads, which takes the output vectors of the previous layer as inputs.

\paragraph{\textbf{Pre-training with E-commerce Corpus}}
During developing the model, our model sometimes generates descriptions with insufficient fluency or inaccurate information due to the limitation of training data. The model never reaches any similar related information for many new products before, making it hard to generate accurate product descriptions. To this end, we introduce domain-specific pre-training into the prefix LM. Instead of utilizing the 
general-domain pre-trained models obtained with large amounts of general knowledge and are not efficient enough for real-time online serving, we pre-train the domain-specific prefix-model with e-commerce knowledge collected from the JD.COM platform. \vspace{-0.4cm}

\begin{figure}[htb]
\centering
\includegraphics[width=0.48\textwidth]{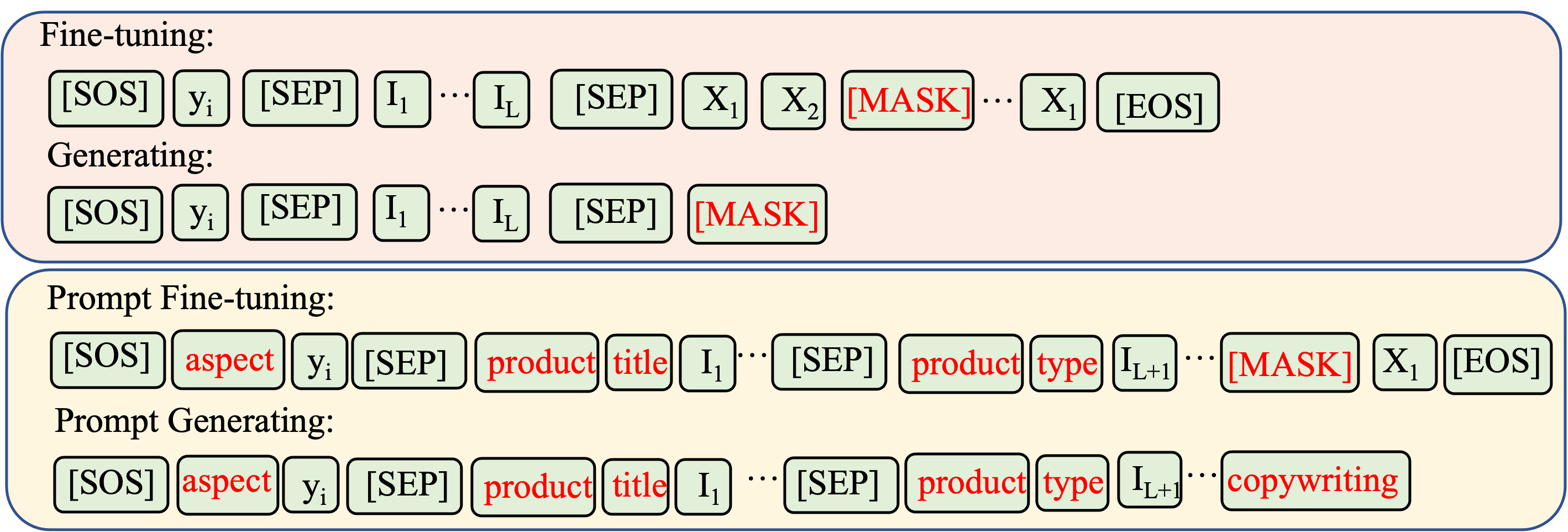}\vspace{-0.2cm}
\caption{\small{The text formulation for fine-tuning and generating process in EPCCG and Prompt-EPCCG}}\vspace{-0.2cm}
\label{fig:input}
\end{figure}\vspace{-0.4cm}

\paragraph{\textbf{Fine-tuning in EPCCG}}
%Specifically, the EPCCG models the distribution $p(w^X_i|w^X_{<i},I,y_i)$ by training on sequences of raw text prepended with control aspects. 
In the fine-tuning process, given the e-commerce pre-trained model, we pack the source sentence by concatenating the name of aspect $y_i$, the relevant product information $I$, and the target copywriting $X$ together with special tokens as ``[SOS]T[SEP]I[SEP]X[EOS]", as shown in Figure~\ref{fig:input}. Here the product information $I$ includes product title, brand, attributes and OCR (optical character recognition) from product advertisement, where each word is initially represented as a vector, as shown in Figure~\ref{fig:frame}.
The model is fine-tuned by masking some percentage of tokens in the target sequence at random, and learning to recover the masked words. The training objective is to maximize the likelihood of masked tokens given context. \footnote{$[EOS]$, which marks the end of the target sequence, can also be masked during fine-tuning, thus when this happens, the model learns when to emit $[EOS]$ to terminate the generation process.}

\paragraph{\textbf{Producing in EPCCG}} In producing process, for each product, we pack the input sentence by combining the product information $I$ with each desired aspect from set $\mathcal{T}$ that is extracted in Section~\ref{sec:aspect ext}. The input sentence is embedded and inputted into EPCCG to predict the remaining sequence as the generated copyrighting.

\subsubsection{Prompt-EPCCG}
% 之前的不论怎样，我们有没有一个场景/情况 即便有classifer 我们也没啥label data for training
% Since very recent year, there is a second sea change in the domain of NLP from the “pre-train, fine-tune” procedure to the “pre-train, prompt, and predict” procedure~\citep{liu2021pre}. 

Recently, a new paradigm to utilize the pre-trained language model called ``prompt-based learning" is proposed to address the issue of fine-tuning the pre-trained model on downstream tasks with few labeled data or no labeled data~\cite{liu2021pre}. In this paradigm, instead of adapting pre-trained LMs to downstream tasks via objective engineering, downstream tasks are reformulated to look more like those solved during the original LM training with the help of a textual prompt. In other words, prompt-based learning is a more efficient way to utilize the knowledge in the pre-trained model with fewer labeled data. Thus, we proposed the ``Prompt-EPCCG" to enhance the performance of EPCCG on personalized copywriting generation when only very few labeled data could be used for training.

% One of the advantages of this method is that, given a suite of appropriate prompts, a single trained LM can be used to solve the downstream tasks with limited number of 
% labeled data but with better performance. 

% Thus, we explored the suitable prompt design for EPCCG and proposed the Prompt-EPCCG for the personalized copywriting generation. 

Specifically, we adopt the Fixed-prompt LM Tuning strategy~\cite{schick2021exploiting} and reorganize the input sequence as 
\begin{align}\nonumber
    &[SOS] aspect: T [SEP] title: I_1 [SEP] brand: I_2 [SEP] attribute: \\\nonumber
    &I_3 [SEP] type: I_4 [SEP] OCR: I_5 [SEP] copywriting: X[EOS]
\end{align}

which means there will be an additional prefix before each detailed product information to help LM better understand the input. We also explore the performance of other possible prompt designs and the experiment results are shown in Section~\ref{sec:exp}.

\subsection{Knowledge-based Post-processing}

At the early stage of development, the generated copywriting still have some issues to be directly displayed into the real-world platform: (1) some mentioned attributes of product in the generated copywriting are not consistent with the real product information; (2) the contents of generated copywriting do not match the desired aspect. To overcome these issues, we introduce the knowledge-based post-process to ensure the quality of the generated copywriting during real-world deployment. The knowledge-based post-process consists of two main steps: attribute-based correction and aspect-based filtering. 

To solve the first issue, we pre-define a list of Regular Expressions (RE) to extract the values of the common attributes from the copywriting. We can get the correct attribute value from our knowledge base. Then we compare the two values and replace the value in copywriting with the correct one. To address the second issue, we resort to the phrase-based aspect classifier mentioned in Section~\ref{sec:aspect_classify}. If the predicted aspect of the generated copywriting does not match the desired one, it will be filtered.  

\section{System Deployment}

\begin{figure}[htb]
\centering
\includegraphics[width=0.35\textwidth]{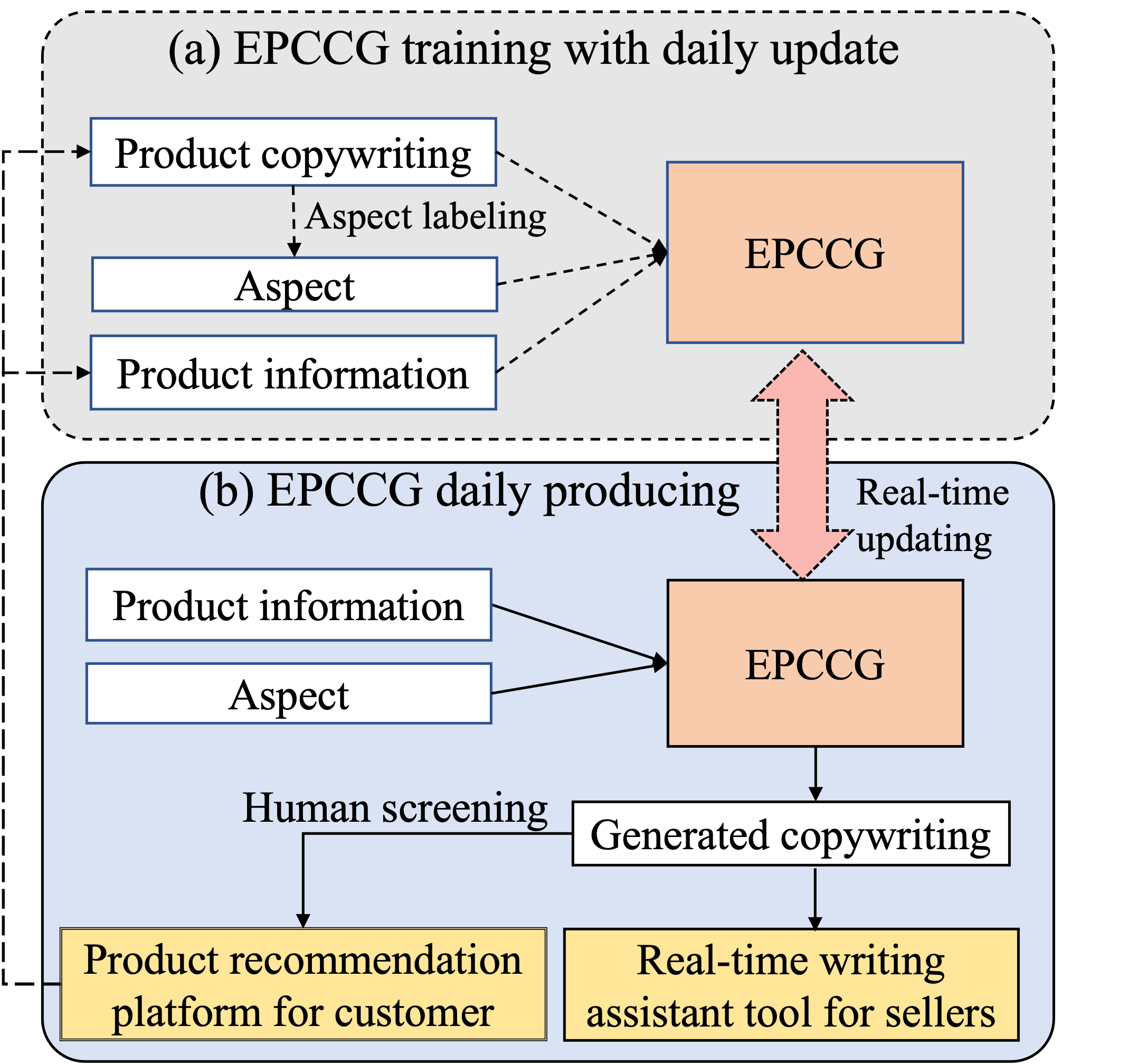} \vspace{-0.1cm}
\caption{\small{Deployment of EPCCG in the E-Commerce recommendation platform}}\vspace{-0.3cm}
\label{fig:deploy}
\end{figure}

In this section, we introduce the experience in how to deploy the proposed EPCCG into our online e-commerce recommendation platform.  As shown in Figure~\ref{fig:deploy}, the overall system consists of the following components:
\begin{itemize}
\item Daily model training: This module collects data, including product information and descriptions, to train the EPCCG model. Each collected product copywriting is assigned an aspect by the phrase-based aspect classifier, as shown in Figure~\ref{fig:deploy} (a).
\item EPCCG daily production: New product information will be collected and combined with each aspect, which will then be inputted into the EPCCG for copywriting generation. After the human screening, the contents produced here are directly used by both the product recommendation platform for JD.com customers.
\item Real-time writing assistant tool for sellers: A real-time writing assistant tool based on updated EPCCG has also been utilized for JD sellers. With this tool, JD sellers can input their desired products, and obtain product descriptions automatically. Sellers can then edit or select the generated contents, and display them to customers.
\item The automatic procedure with daily update: EPCCG is trained daily to meet the latest writing preference based newly updated data collected from the recommendation platforms.
\end{itemize}

\section{Experiment and Payoff}
\label{sec:exp}

In this section, we first introduce the dataset, the compared models and the evaluation metrics. We then demonstrate the experimental results in a series of evaluations and further analyze the development selections. In addition, the real industry payoff of the proposed EPCCG system after deployment and several practical cases are provided. Details of implementation are provided in Appendix~\ref{app:parameter}.   

\vspace{-0.2cm}
\begin{table}[htb]
\begin{center}
\caption{The extracted aspects of different product categories}\vspace{-0.3cm}
\label{tab:dataset_aspect}
\begin{tabular}{ll}
\toprule
Dataset                    & Extracted aspects \\\midrule
% Mobile-phone& appearance,screen, network,camera\\
% &battery,security,capability \\\hline
 computer& appearance,screen, graphic processing\\
 &battery,security, heat dissipation, keyboard\\
 &capability, camera\\\hline
men clothes &fabric,version, pattern, color, match, style, \\
&pocket, collar                     \\\hline
%women clothes &fabric,version, pattern, color, match, style, \\
%&pocket, collar                     \\\hline
%facial skin care  & sun-screening, brightening, moisturizing, \\
%&cleaning, anti-wrinkling, package design \\
%&repair and soothing, ingredient \\\hline
home appliances & performance, degerming, capability, noises \\
&intelligent control, energy conservation\\
&appearance\\
 \bottomrule
\end{tabular}
\end{center}
\end{table}
\vspace{-0.2cm}

\small
\begin{table*}[htb]
\begin{center}
\caption{Machine-based and human-based evaluation results on mobile datasets.}\vspace{-0.2cm}
\label{tab:comparison result}
\begin{tabular}{llllllllllll}
\toprule
\textit{methods} &\textit{rouge-1}&\textit{rouge-2}&\textit{rouge-L}&\textit{bleu-1}&\textit{bleu-2}&\textit{bleu-3}&\textit{bleu-4}&\textit{meteor}&\textit{sacrebleu}&\textit{aspect}&\textit{validness}\\\midrule
CTRL&0.2056&0.0954&0.1723&0.1637&0.1109&0.0806&0.0627&0.1603&3.9511&91.05\%& 93.88\%\\\hline
 conditional GPT&0.2012&0.0667&0.1917&\textbf{0.2247}&0.1305&0.0935&	\textbf{0.0938}&0.1338&5.0886&95.55\% &95.66\%         \\\hline
EPCCG w/o pretraining &0.2023&0.0635&0.1875&0.1886&0.1013&0.1009&0.0947&0.1394&5.2836&	95.00\% &95.80\% \\\hline
 discrete-code based EPCCG&0.2228&0.1072&0.1877&0.1770&0.1235&0.0918&0.0725&0.1680&4.6739&94.32\% &96.25\% \\\hline
 label-code based EPCCG&0.2190&0.1073&0.1836&0.1669&0.1179&0.0883&0.0699&0.1728&4.7519&93.02\% &96.10\% \\\hline
 name-code based EPCCG&\textbf{0.2230}&\textbf{0.1234}&\textbf{0.2105}&0.1909&\textbf{0.1359}&\textbf{0.1027}&0.0821&\textbf{0.1770}&	\textbf{5.9966}&\textbf{95.71}\%&\textbf{96.29\%} \\
 \bottomrule
\end{tabular}
\end{center}
\end{table*}
\normalsize

\subsection{Dataset}

Considering that there is a lack of large-scale open-source datasets for this task, we constructed a new dataset \textit{JDCopywriting}, containing the basic information of the products, including title, OCR text, attribute, as well as the copywriting of products with labeled aspects. The data are collected from JD.COM platform, a large-scale website for e-commerce in China.
The product information and the copywriting are composed by the sellers and content writers on the website from Jan 2020 to June 2021. Each data instance is automatically annotated with its aspect based on the proposed phrase-based aspect classifier introduced in Section~\ref{sec:aspect_classify}. 

The whole dataset covers six categories of products, including \textit{men clothes}, \textit{mobile-phone}, \textit{computers}, \textit{women clothes}, \textit{facial skin care}, and \textit{home appliances}, each of which covers $500,000$ data samples in total. Each of them are extracted with several aspects, as illustrated in Table~\ref{tab:dataset_aspect}. The aspects of more categories are provided in Appendix~\ref{app:aspect}.

\subsection{System for Comparison}
In this section, we introduce the baseline and choices for our model components while developing EPCCG and the prompt-EPCCG.

\subsubsection{Baseline of EPCCG}
\label{sec:baseline}

To validate the superiority of the proposed EPCCG, we compare it with two SOTA controllable text generation models: \vspace{-0.1cm}
\begin{enumerate}
    \item \textbf{Conditional Transformer Language Model(CTLR)}~\cite{keskar2019ctrl} which is a conditional transformer based LM, trained to condition on control codes that govern style, content, and task-specific behavior.
    \item \textbf{Conditional GPT}~\citep{kieuvongngam2020automatic} which is based on OpenAI GPT-2 to generate abstractive and comprehensive information based on keywords extracted from the original articles. 
\end{enumerate}

\subsubsection{Choices of Controllable Code Pattern}
\label{sec:choices_of_code}
To explore the best way to involve the controllable aspect in the input sentence, we test three settings:\vspace{-0.1cm}
\begin{enumerate}
    \item \textbf{Discrete-code based} treats the controllable aspect as a category label by a one-hot vector. The one-hot vector is then concatenated with the input product information.
    \item \textbf{Label-code based} also treats the controllable aspect as a category label but tokenizes it as a special token.
    \item  \textbf{Name-code based} directly concatenates the name of the aspect with the product information for tokenization.
\end{enumerate} 

\subsubsection{Choices of Prompt Design}
\label{sec:prompt_result}

The key point to a successful prompt-model is the design of the prompt paradigm. To explore the best prompt paradigm, we compare the five typical settings. \vspace{-0.1cm}
\begin{enumerate}
 %   \item  \textbf{prompt-EPCCG w/o finetune} where the pre-training model is fixed and directly generate copywriting with the prompt format of input. 
    \item \textbf{prompt-EPCCG w/o sep}, where the prefix is added in a simple way such as ```\textit{aspect:...product:...copywriting:...}" to format the input.
    \item \textbf{prompt-EPCCG with sep}, where we add the prefix with ```\textit{[SEP]}'' token as ```\textit{aspect:....[SEP] product:....[SEP] copywriting:...}" to format the input.
    \item \textbf{prompt-EPCCG-advance w/o sep}, where we further separate the input with more detailed prefix as ```\textit{aspect:...product title:...product type:...product attribute:...copywriting:...}".
    \item \textbf{prompt-EPCCG-advance with sep} add ```\textit{[SEP]}''  onto the input of \textbf{prompt-EPCCG-advance w/o sep}.
\end{enumerate}

\subsection{Evaluation Metrics}
We evaluate our model on generation quality and aspect capturing ability based on human and machine evaluation metrics.

\textbf{Machine-based Evaluation for Copywriting Quality}. A series of typical BLEU (i.e., bleu-1, bleu-2, bleu-3, bleu-4 and sacrebleu) as well as ROUGE scores (i.e., rouge-1, rouge-2, rouge-L and meteor) are utilized to evaluate the similarity between the generated and ground-truth copywriting regarding the N-gram cases. 

\textbf{Machine-based Evaluation for Aspect Capturing Ability}. To judge whether the generated product copywriting matches the desired controllable aspect, we utilize the phrase-based aspect classifier\footnote{The well-trained classifier is obtained and stored during the data labeling process as introduced in Section~\ref{sec:aspect_classify}.} to classify the aspect of the generated copywriting. We calculate the percentage of the correctly matched copywriting among the total generation, mentioned as \textit{aspect}. 

\textbf{Human-based Evaluation}. We also depend on human reviewers to judge whether a produced copywriting is valid and informative for online display, mentioned as \textit{validness}. It is worth noting that this evaluation is also an online metric since deployment.  

\subsection{Experiment Results and Analysis}

In this section, we analyze the experimental results by focusing on a few issues illustrated as below. We utilize the \textit{men clothes} category as an example to explore the answers to the issues. 

\begin{table*}[htb]
\begin{center}
\caption{Machine-based and human-based evaluation results on different prompt setting for EPCCG.}\vspace{-0.2cm}
\label{tab:prompt result}
\begin{tabular}{lllllllllll}
\toprule
\textit{datasests} &\textit{rouge-1}&\textit{rouge-2}&\textit{rouge-L}&\textit{bleu-1}&\textit{bleu-2}&\textit{bleu-3}&\textit{bleu-4}&\textit{meteor}&\textit{sacrebleu}&\textit{aspect}\\\midrule
%prompt-EPCCG w/o finetune&\\\hline
prompt-EPCCG w/o sep &0.2282&0.0755&0.2114&0.3004&0.2085&0.1492&0.1131&0.1437&8.2289&96.77\%         \\\hline
prompt-EPCCG with sep &0.2293&0.0755&0.2125&\textbf{0.3019}&\textbf{0.2092}
&0.1492&0.1133&\textbf{0.1444}&8.3290&97.22\% \\\hline
prompt-EPCCG-advance w/o sep&0.2275&0.0754&0.2107&
0.2959&0.2056&0.1472&0.1114&0.1422&8.2096&97.02\%\\\hline
prompt-EPCCG-advance with sep & \textbf{0.2317}&\textbf{0.0791}&
\textbf{0.2157}&0.2966&0.2072&\textbf{0.1495}&\textbf{0.1141}&0.1432&\textbf{8.51}77&\textbf{97.42\%}\\
 \bottomrule
\end{tabular}
\end{center}
\end{table*}

\begin{table*}[htb]
    \centering
     \caption{The produced copywriting generated by varying the aspect attribute while fixing the product title as input}\vspace{-0.2cm}
    \begin{tabular}{l|l|l}
    \toprule
        category&aspects & produced copywriting \\\hline
         computers & ``appearances" &``This notebook uses a thin and light body design, easy to carry, more handy when going\\ 
         & & out to work"\\ \cline{2-3}
         &``heat dissipation" &``Built-in high-density thin fan blades and double heat pipe design, the heat is discharged \\ 
          & & through the whole cooling system".\\ \cline{2-3}
         &``battery"&``The built-in large-capacity battery has a long battery life, helping you to bid farewell to \\
        &&the power crisis calmly".\\\hline
        home appliances & ``performance" &``Equipped with D-type evaporator to increase the cooling area and improve the cooling \\
        &&performance".\\\cline{2-3}
        &``noises"&``Using Fisher-Paykel direct drive variable frequency motor, the operation is smooth and \\
        &&smooth, and the noise is effectively reduced".\\\cline{2-3}
        & ``energy conservation" & ``First-class energy efficiency, providing energy saving and electricity saving".\\
    \bottomrule
    \end{tabular}
    \label{tab:cases}
\end{table*}

\vspace{0.2cm}
\noindent\textbf{Does different pattern of control code influence the performance of copywriting generation?}
We compared three different patterns of control aspect involvement as mentioned in Section~\ref{sec:choices_of_code}. The experiment results on \textit{men clothes} category are provided in Table~\ref{tab:comparison result}. Based on the results, concatenating the name of aspect with the input text and tokenizing it in the same way as the input product information achieved the best performance. Specifically, the \textit{name-code based EPCCG} outperforms the other two patterns with an advantage of $+1.34$ sacreBLEU (relatively $21.2\%$). The highest \textit{aspect} of \textit{name-code based EPCCG} validates that involving the semantic meaning of aspect help improve the aspect capturing capability of the text generation model. Thus, we finally select the \textit{name-code based EPCCG} as the final implemented pattern.  

\vspace{0.2cm}
\noindent\textbf{Does the proposed EPCCG architecture bring the best performance for copywriting generation?} We compared the proposed EPCCG with the state-of-the-art controllable text generation model as mentioned in Section~\ref{sec:baseline}. As shown in Table~\ref{tab:comparison result}, the \textit{name-code based EPCCG} achieves a higher score in seven ROUGE and BLEU metrics over CTRL and conditional GPT. Whichever controllable code pattern is selected, the proposed EPCCG have the highest validness by human evaluation in the real-world industrial situation over CTRL and conditional GPT by $2.42\%$
and $0.63\%$. %The results that \textit{name-code based EPCCG} outperforms the \textit{EPCCG w/o pretraining} with an advantage of $+0.5$ \textit{aspect} demonstrates the necessity of e-commerce pre-trained model.

\vspace{0.2cm}
\noindent\textbf{Is the proposed prefix-based pre-training strategy necessary for the improvement of performance?} We conduct an ablation study by comparing the performance of the proposed EPCCG with and without the pre-training strategy. As shown in Table~\ref{tab:comparison result}, given the \textit{name-code based EPCCG} and \textit{EPCCG w/o pre-training} setting which have the same architecture and controllable code pattern, \textit{name-code based EPCCG} significantly outperforms \textit{EPCCG w/o pre-training} in almost all the evaluation metrics, such as an advantage of $+0.5$ \textit{aspect}. Specially, adding domain-specific pre-training strategy improved the \textit{meteor} by around $4.32$.

\vspace{0.2cm}
\noindent\textbf{Does the design of the prompt paradigm influences the quality of the generated copywriting?} As shown in Table~\ref{tab:prompt result}, we compare the different designs of prompt paradigm mentioned in Section~\ref{sec:prompt_result} based on the \textit{men clothes} category. Overall, the setting \textit{prompt-EPCCG with sep} achieves the best performance regarding both the quality of generated copywriting as well as the capturing ability of aspect. Based on the results that \textit{prompt-EPCCG with sep} outperforms \textit{prompt-EPCCG w/p sep} with advantage of $+0.45$ aspects and \textit{prompt-EPCCG-advance with sep} outperforms \textit{prompt-EPCCG-advance w/o sep} with advantage of $+0.38$ aspects, it can be conclude that adding the token of ``[SEP]" improves the aspect capturing ability. Based on the results that \textit{prompt-EPCCG-advance with sep} outperforms \textit{prompt-EPCCG with sep} with advantage of $+0.2$ aspects and $+0.2$ sacreBLEU, it can be concluded that formatting more structured product information text can help the model better understand the semantic meaning of input. %By comparing with the base EPCCG results in Table~\ref{tab:comparison result}, prompt-EPCCG sucesfully lift the \textit{aspect} by $1.71$ and \textit{sacreBLEU} by $2.5$. 

\subsection{Case Studies}

In this section, we perform case studies to observe how the proposed EPCCG influenced the generation so that the model can generate different contents based on various controllable aspects. Table~\ref{tab:cases} presents the generated examples by the models with varying aspects for home appliances and computers (have been translated into English). The generated copywriting can talk about the product's different aspects by successfully capturing the desired.

\begin{figure}[htb]
\centering
\includegraphics[width=0.98\columnwidth]{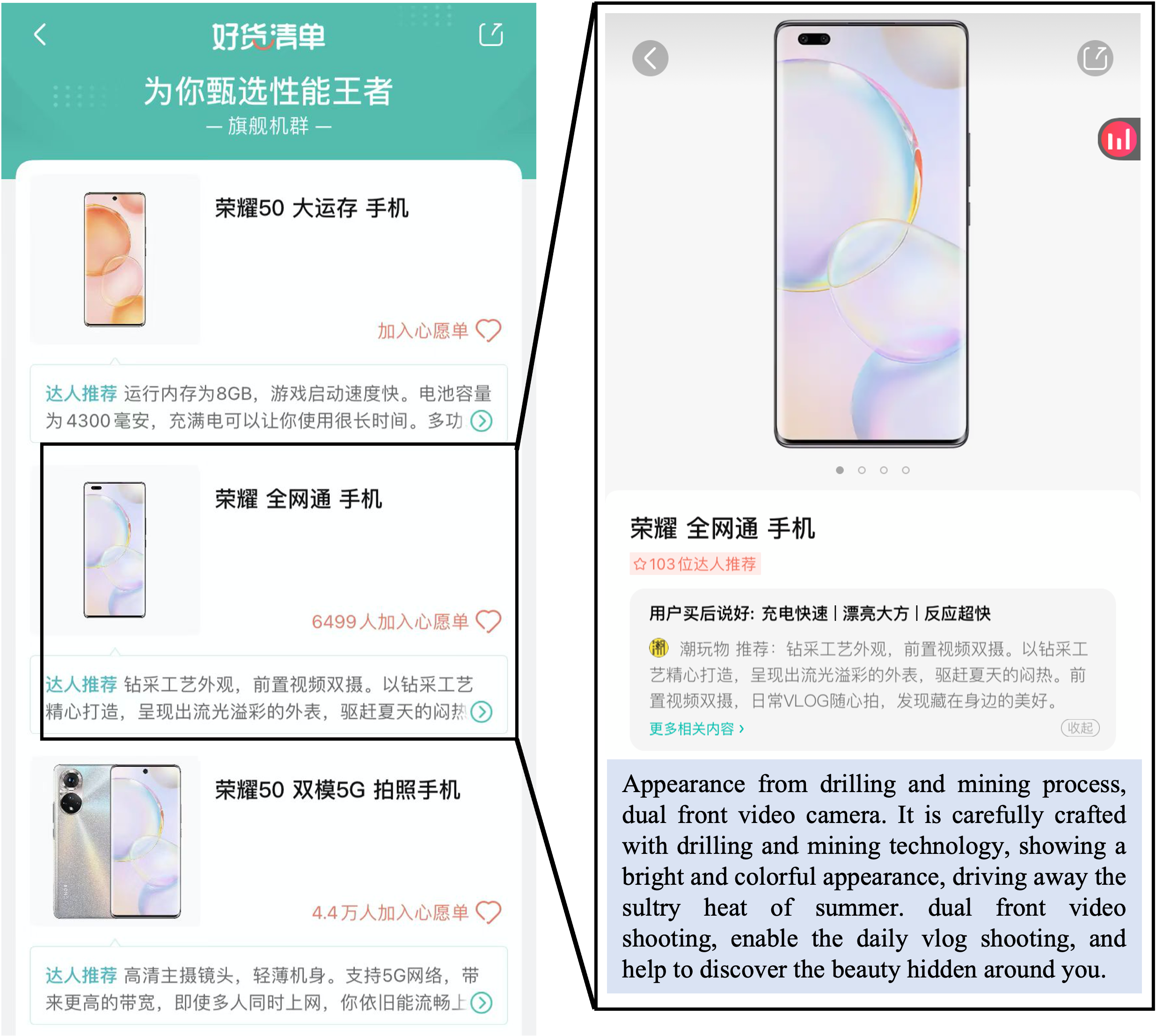} 
\caption{Examples of EPCCG generated descriptions on the JD e-commerce product recommendation platform.}\vspace{-0.2cm}
\label{fig4}
\end{figure}\vspace{-0.2cm}

\subsection{Payoff After Deployment}

EPCCG has been deployed in the JD.com product recommendation platform since October 2021. Since deployment, EPCCG has covered 6 main categories of products, including women clothes, men clothes, mobile-phone, computers, appliances, and facial care, by generating and delivering \textbf{56,000} copywriting and promoting \textbf{millions of Gross Merchandise Volume}. The deployed EPCCG system serves for the content generation for customers. Figure~\ref{fig4} shows an example of the product recommendation platform for customers, “Discovery Goods Channel”. The English translations of the generated texts are shown in the pop up bubbles in the figure. The generated product descriptions, paired with a short title, are pushed to the Discovery Goods Channel in the JD e-commerce website.

\section{Conclusions}
In this paper, we introduced a deployed controllable product copywriting generation system in a large-scale e-commerce platform. We proposed a novel model EPCCG and its extension prompt-EPCCG based on the prefix-based pre-trained language model. Our extensive experiments showed that our method outperformed the baseline models through various evaluation metrics, including machine-based and human evaluations. We have successfully deployed the proposed framework onto JD.com, one of the world’s largest online e-commerce platforms, with significantly payoff. A large volume dataset for product description generation, namely JDCopywriting, has been generated and annotated, which can be used as a benchmark dataset for future research in this field.

\bibliographystyle{ACM-Reference-Format}
\bibliography{main}

%%
%% If your work has an appendix, this is the place to put it.

\appendix

\section{Hyper-parameters and architecture for reproducing}
\label{app:parameter}

\subsection{The hyper-parameters of phrase-based aspect classifier}

In this section, we introduce the detailed setting of the proposed phrase-based aspect classifier. To select the substitute words for each aspect, we select the Top-50 words which with the highest frequent appearances, namely we let $K=50$. Next, the core components of the proposed classifier is the pre-trained language model. The pre-trained model is based on BERT model which consists of $12$ transformer layers. While training, the learning rate is $0.00003$ and the batch size is $220$. The detailed hyper-parameters for architecture are listed in Table~\ref{tab:app_para1}.

\begin{table}[htb]
    \centering
    \caption{The detailed hyper-paramters of arechtecture of pre-trained LM in aspect classifier.}
    \begin{tabular}{l|l}
    \toprule
        \textbf{hyper-parameters} &\textbf{value}\\\hline
         attention\_probs\_dropout\_prob&0.1\\\hline
        embedding\_size&768\\\hline
        hidden\_act&"gelu"\\\hline
        hidden\_dropout\_prob& 0.1\\\hline
        hidden\_size&192\\\hline
        initializer\_range&0.02\\\hline
        intermediate\_size&768\\\hline
        layer\_norm\_eps&1e-12\\\hline
        max\_position\_embeddings&512\\\hline
        num\_attention\_heads& 3\\\hline
        num\_hidden\_layers& 12\\\hline
        output\_past& true\\\hline
        pad\_token\_id& 0\\\hline
        summary\_activation& "gelu"\\\hline
        summary\_last\_dropout& 0.1\\\hline
        type\_vocab\_size& 2\\\hline
        vocab\_size&21128\\
        \bottomrule
    \end{tabular}
    \label{tab:app_para1}
\end{table}

\subsection{The hyper-parameters of EPCCG model}
In this section, we introduce the hyper-parameters of the architecture of prefix-based controllable copywriting generation model while implementation. The backbone network is a 12-layer transformer with multi-head attentions. While training, the learning rate is $0.00003$ and the batch size is $220$. The detailed hyper-parameters for architecture are listed in Table~\ref{tab:app_para2}.

\begin{table}[htb]
    \centering
    \caption{The detailed hyper-paramters of architecture of pre-trained LM in aspect classifier.}
    \begin{tabular}{l|l}
    \toprule
        \textbf{hyper-parameters} &\textbf{value}\\\hline
         attention\_probs\_dropout\_prob&0.1\\\hline
        embedding\_size&768\\\hline
        hidden\_act&"gelu"\\\hline
        hidden\_dropout\_prob& 0.1\\\hline
        hidden\_size&192\\\hline
        initializer\_range&0.02\\\hline
        intermediate\_size&768\\\hline
        layer\_norm\_eps&1e-12\\\hline
        max\_position\_embeddings&512\\\hline
        num\_attention\_heads& 12\\\hline
        num\_hidden\_layers& 12\\\hline
        output\_past& true\\\hline
        pad\_token\_id& 0\\\hline
        summary\_activation& "gelu"\\\hline
        summary\_last\_dropout& 0.1\\\hline
        type\_vocab\_size& 2\\\hline
        vocab\_size&21128\\
        \bottomrule
    \end{tabular}
    \label{tab:app_para2}
\end{table}

\small
\begin{table*}[!t]
\begin{center}
\caption{Machine-based and human-based evaluation results on different categories of aspects based EPCCG.}\vspace{-0.2cm}
\label{tab:datasets result}
\begin{tabular}{llllllllllll}
\toprule
\textit{datasests} &\textit{rouge-1}&\textit{rouge-2}&\textit{rouge-L}&\textit{bleu-1}&\textit{bleu-2}&\textit{bleu-3}&\textit{bleu-4}&\textit{meteor}&\textit{sacrebleu}&\textit{aspect}&\textit{validness}\\\midrule
mobile phone&0.2054&	0.0674&0.1904&0.2956&0.2015&0.1432&0.1082&	0.1386&7.6870&96.02\%&96.29\%\\\hline
women clothes&0.1770&0.0454&0.1641&0.2584&0.1669&0.126&	0.0938&0.1338&5.0886&95.17\% &91.40\%         \\\hline
facial skin care&0.1863&0.0468&0.1667&0.2581&0.1702&0.1138&	0.0810&0.1429&5.5697&90.50\%&96.25\% \\\hline
home appliances&0.1928&0.0606&0.1825&0.2660&0.1804&0.1262&0.0936&	0.1266&6.2596&97.10\%&99.45\%\\\hline
computer& 0.1969&0.0653	&0.1811	&0.2112&0.1513&0.1110&0.0849&	0.1684&7.1225&96.16\%&95.06\%\\
 \bottomrule
\end{tabular}
\end{center}
\end{table*}\normalsize

\section{Preliminary of LDA}
\label{app:lda}
Topic Modelling in natural language document aims to use unsupervised learning to extract the main topics (represented as a set of words) that occur in a collection of documents.  LDA~\citep{jelodar2019latent} is an unsupervised generative probabilistic method for modeling a corpus, which is the most commonly used topic modeling method. It assumes that each document can be represented as a probabilistic distribution over latent topics, and that topic distribution in all documents share a common Dirichlet prior. Each latent topic in the LDA model is also represented as a probabilistic distribution over words and the word distributions of topics share a common Dirichlet prior as well. In the process of product aspect extraction, given the copywriting dataset $\mathcal{D}=\{x_i\}^{N}_{i=1}$ and each copywriting $x_i$ contains $H^{(i)}$ words $w_k^{(i)}$, the whole copywriting corpus $\mathcal{D}$ is modeled in the following process:
  \begin{itemize}
      \item  Choose a distribution $\phi \backsim Dir(\beta)$ for aspect.
      \item Choose a distribution $\theta \backsim Dir(\alpha)$ for copywriting.
      \item For each word $w_k^{(i)}$ in copywriting $x_i$:\\
          (a) Choose a topic $t_m^{(i)} \backsim Multinomial(\phi)$.\\
          (b) Choose a word $w_k^{(i)}\backsim Multinomial(\theta)$.
  \end{itemize}
Thus, to learn this generative mode, the goal is to maximize the overall objective as:
\begin{equation}
    \max_{\phi, \theta} \prod^N_{i=1}\int p(\theta^{(i)}|\alpha)(\prod^{H^{(i)}}_{k=1}\sum_{t_m^{(i)}}p(z_m^{(i)}|\theta^{(i)})p(w_k^{(i)}|t_m^{(i)},\beta)))d\theta^{(i)}
\end{equation}
In the process of product aspect extraction, given the copywriting dataset $\mathcal{D}=\{x_i\}^{N}_{i=1}$ and each copywriting $x_i$ contains $H^{(i)}$ words $w_k^{(i)}$.
where $\alpha$, $\beta$ and $M$ are hyper-parameters about LDA. Here We use the Gibbs sampling, which is a Monte Carlo Markov-chain algorithm to estimate the LDA parameters. After this, we can successfully identify the main aspects from the copywriting corpus.

\section{Preliminary: Prefix Language Model}
\label{sec:pre_LM}

The prefix Language Model (LM)~\cite{dong2019unified} is a left-to-right Language Model that decodes output sentence on condition of a prefixed input sequence, which is encoded by the same model parameters with a bi-directional mask. As
shown in Figure 4, during training process, the input tokens can attend to each other bidirectionally, while the output tokens can only attend to tokens on the left. To realize this, a corrupted text reconstruction objective is usually applied over input tokens, and an auto regressive language modeling objective is applied over output tokens, which encourages the prefix LM to better learn representations of the input. In this way, the prefix LM can model the generative process of $p(w^X_i|w^X_{<i})$ for each sentences.

Specifically, the input vectors is first packed as $H^0 = [ w^{I_1}, . . . , w^I_{L_1}]$, and then encoded into contextual representations at different levels of abstract $H^l =[h^{l_t},h^{l_1},···,h^{l_n}]$ using an $L-th$ layer Transformer $H^l=\textit{Transformer}^l(H^{(l-1)}), l \in [1, L]$. In each Transformer block, multiple self-attention heads are used to aggregate the output vectors of the previous layer. For the $l-th$ Transformer layer, the output of a self-attention head $A^l$ is computed via:

\begin{align}
 Q=H^{l-1}W_Q^l, K=H^{l-1}W_{K}^l, V=H^{l-1}W_V^l\\
 M_{ij}=\begin{cases}
 0& \textit{allow to attend}\\
 -\infty &\textit{prevent from attending}
 \end{cases}\\
 A_l=softmax(\frac{QK^\top}{\sqrt{d_k}})+M)V_l
\end{align}

where the previous layer’s output, namely, $H^{(l-1)} \in R^{n\times d_h}$, is linearly projected to a triple of queries, keys and values using parameter matrices $W^l_Q$ , $W^l_K$ , $W^l_V \in R^{d_h\times d_k}$, respectively, and the mask matrix $M \in R^{n\times n}$ determines whether a pair of tokens can be attended to each other.

\section{The extracted aspects for different categories of products}
\label{app:aspect}

In this section, we provide the extracted aspects for all the six categories of products in the JDCopywriting dataset.

\begin{table}[htb]
\begin{center}
\caption{The extracted aspects for different categories of products.}
\label{tab:dataset}
\begin{tabular}{ll}
\toprule
Dataset & Extracted aspects \\\midrule
 Mobile-phone& appearance,screen, network,camera\\
 &battery,security,capability \\\hline
 computer& appearance,screen, graphic processing\\
 &battery,security, heat dissipation, keyboard\\
 &capability, camera\\\hline
men clothes &fabric,version, pattern, color, match, style, \\
&pocket, collar                     \\\hline
women clothes &fabric,version, pattern, color, match, style, \\
&pocket, collar                     \\\hline
facial skin care  & sun-screening, brightening, moisturizing, \\
&cleaning, anti-wrinkling, package design \\
&repair and soothing, ingredient \\\hline
home appliances & performance, degerming, capability, noises \\
&intelligent control, energy conservation\\
&appearance\\
 \bottomrule
\end{tabular}
\end{center}
\end{table}

\section{Evaluation results on different datasets}

As shown in Table\ref{tab:datasets result}, the proposed EPCCG shown effectiveness in the other five categories of products. Specifically, the proposed EPCCG performs the best while dealing with the category of \textit{home appliance}. This is because the  \textit{home appliance} has the most standard attributes information which make the input text much more structured. Following \textit{home appliance} are \textit{mobile phones} and \textit{computers}, which are all belong to the electrical line and have the most standard attribute information, resulting the standard aspects.

\end{document}